\documentclass[sigconf]{acmart}

\usepackage{booktabs} % For formal tables
\usepackage{subcaption}

\usepackage[english]{babel}
\usepackage[utf8x]{inputenc}
\usepackage[T1]{fontenc}
\usepackage{ragged2e}

\usepackage{mathtools}
\usepackage{bm}

\usepackage{amsmath}
\usepackage{graphicx}
\usepackage[colorinlistoftodos]{todonotes}

\newcommand{\el}{et al.}

\copyrightyear{2020}
\acmYear{2020}
\setcopyright{acmcopyright}
\acmConference[GECCO '20]{Genetic and Evolutionary Computation Conference}{July 8--12, 2020}{Cancún, Mexico}
\acmBooktitle{Genetic and Evolutionary Computation Conference (GECCO '20), July 8--12, 2020, Cancún, Mexico}
\acmPrice{15.00}
\acmDOI{10.1145/3377930.3389821}
\acmISBN{978-1-4503-7128-5/20/07}

\begin{document}
\title[Interactive Evolution Within Latent Level-Design Space of GANs]{Interactive Evolution and Exploration Within Latent Level-Design Space of Generative Adversarial Networks}
%\titlenote{Produces the permission block, and copyright information}
%\subtitle{Subtitle}
%\subtitlenote{The full version of the author's guide is available as
%  \texttt{acmart.pdf} document}

%%% The submitted version for review should be ANONYMOUS

\author{Jacob Schrum}
\orcid{0000-0002-7315-0515}
\affiliation{%
  \institution{Southwestern University}
  \streetaddress{1001 E. University Ave}
  \city{Georgetown} 
  \state{Texas, USA} 
  \postcode{78626}
}
\email{schrum2@southwestern.edu}

%%% My undergraduate student who did a lot of the Zelda work
\author{Jake Gutierrez}
\affiliation{%
  \institution{Southwestern University}
  \streetaddress{1001 E. University Ave}
  \city{Georgetown} 
  \state{Texas, USA} 
  \postcode{78626}
}
\email{gutierr8@southwestern.edu}

\author{Vanessa Volz}
\affiliation{%
  \institution{modl.ai}
  \streetaddress{Nørrebrogade 184, 1}
  \city{Copenhagen, Denmark} 
}
\email{vanessa@modl.ai}

\author{Jialin Liu}
\affiliation{Southern University of Science and Technology
\city{Shenzhen, China}
}
\email{liujl@sustech.edu.cn}

\author{Simon Lucas}
\affiliation{
    \institution{Queen Mary University of London}
    \city{London, UK}
}
\email{simon.lucas@qmul.ac.uk}

\author{Sebastian Risi}
\affiliation{%
  \institution{modl.ai, IT University of Copenhagen}
  \streetaddress{Nørrebrogade 184, 1}
  \city{Copenhagen, Denmark} 
}
\email{sebastian@modl.ai}

\begin{abstract}
Generative Adversarial Networks (GANs) are an emerging form of indirect encoding. The GAN is trained to induce a latent space on training data, and a real-valued evolutionary algorithm can search that latent space. Such Latent Variable Evolution (LVE) has recently been applied to game levels. However, it is hard for objective scores to capture level features that are appealing to players. Therefore, this paper introduces a tool for interactive LVE of tile-based levels for games. The tool also allows for direct exploration of the latent dimensions, and allows users to play discovered levels. The tool works for a variety of GAN models trained for both \emph{Super Mario Bros.} and \emph{The Legend of Zelda}, and is easily generalizable to other games. A user study shows that both the evolution and latent space exploration features are appreciated, with a slight preference for direct exploration, but combining these features allows users to discover even better levels. User feedback also indicates how this system could eventually grow into a commercial design tool, with the addition of a few enhancements.
\end{abstract}





\maketitle

\section{Introduction}

Recent work \cite{volz:gecco2018,giacomello:cog19,park:cog19,torrado2019bootstrapping,gutierrez2020zeldagan} has shown that it is possible to learn the structure of video game levels using Generative Adversarial Networks (GANs~\cite{goodfellow2014generative}). Although GANs can learn a representation of the underlying structure, the semantics of the game are generally not included in the representation. As a result, levels may not be beatable, because there may not be a key to a door, or a path between two points of interest. To combat this issue, some GAN-based approaches use some form of bootstrapping to enhance solvability \cite{park:cog19,torrado2019bootstrapping}. Similarly, the original work applying GANs to Mario \cite{volz:gecco2018} used latent variable evolution (LVE) to favor levels that could actually be beaten as determined by AI simulations. However, fitness-based optimization tends to converge on a limited set of points within a search space, which ignores evolution's massive potential for exploration.

In addition, it is often difficult to formally define 
an evaluation function that can assess
the quality of generated game levels automatically. Even with divergent search techniques such as Novelty Search \cite{lehman:ecj2011} and MAP-Elites \cite{mouret:arxiv15},
which seek to explore large areas of the search space, some explicit characterization of the space is needed (behavior characterization or binning scheme) to dictate which offspring are preferred.

An alternative is interactive evolution, which can explore a search space with a human in the loop \cite{eiben:interactive2015,bontrager2018deep,zaltron2019cg,takagi2001interactive}. Interactive evolution allows a user to select whatever options are most appealing at the moment, allowing for serendipitous discovery of novel solutions without formalized domain knowledge. This paper applies interactive latent variable evolution to GANs trained on two video games: Super Mario Bros.\ and The Legend of Zelda. 

%The benefit of interactive evolution is the instant feedback of seeing the evolving population, and the chance to adjust the direction of the search dynamically.
A typical problem with interactive evolution, however, is a user's frustration with evaluating many low-quality individuals. 
Using a GAN as the mapping function addresses this issue \cite{bontrager2018deep,zaltron2019cg} by assuring that most genotypes presented to the user lead to well-formed phenotypes.

%In this paper, we suggest to address this issue by assuring that most genotypes presented to the user lead to well-formed phenotypes by using a GAN as the mapping function \cite{bontrager2018deep,zaltron2019cg}.

Besides standard interactive evolution, the system provides users a rich environment for interaction with the generated content. For example, users are also able to directly edit latent vectors, providing another means of latent space exploration. Applying the search to game levels also means that users can immediately play their creations, and although some levels are unbeatable, the overall experience is an enjoyable game in itself. The resulting software is available online\footnote{\url{https://github.com/schrum2/GameGAN}} 
and can be used to interactively evolve Mario and Zelda levels without modification. The approach is also easily extensible, and should therefore allow for interactive latent space exploration in various other domains.

A small user study is conducted to assess the system, and gather suggestions for improvement. In particular, users searched latent space using only evolution, or only direct latent vector editing. After trying both, they indicated their preference. Users then tried the combined system and were asked if the combination made results even better. Users had a slight preference for direct exploration, but found that combining search techniques was better than either individually. Users also gave advice for improving the system.

%% If we are running out of space, this paragraph will be the first thing to go
The paper proceeds by discussing related work (Section \ref{sec:related}) and the particular video game domains used in this paper: Mario and Zelda (Section \ref{sec:domains}). Then the approach introduced in this paper is described (Section \ref{sec:approach}). Next, the procedure and results for a human subject study designed to analyse the new system are presented (Section \ref{sec:humanStudy}). Finally, the results and possible future work are discussed (Section \ref{sec:discussion}) before concluding (Section \ref{sec:conclusion}).

\section{Related Work}
\label{sec:related}
Procedural Content Generation for games is discussed, followed by descriptions of technical tools used in this paper such as Generative Adversarial Networks (GANs), which are used as a genotype-to-phenoype mapping for interactive evolution of game levels. 

\subsection{Procedural Content Generation}

AI methods have been widely used for generating game content (e.g.\ game rules, levels, maps/mazes, characters, items, vehicles, stories, textures, and sound) with limited or indirect user input~\cite{hendrikx2013procedural,shaker2016procedural,yannakakis2018artificial}. Yannakakis and Togelius~\cite{yannakakis2018artificial} divide existing Procedural Content Generation (PCG) methods into the following categories: search-based methods, solver-based methods, grammar-based methods, cellular automata, and noise/fractals, all of which have been applied to the automatic generation of game levels/dungeons. 
In recent years, training of machine learning models on existing game contents to generate new content has emerged as a viable approach~\cite{summerville2018procedural}: PCG via Machine Learning (PCGML).
Two popular domains for applying PCGML are the generation of Mario levels and dungeons.

Dungeons are a common type of game level. The earliest algorithmic creation of dungeons goes back to Rogue (1980)~\cite{yannakakis2018artificial}. Since then, many PCG methods have been used to generate dungeons~\cite{adams2002automatic,Linden2014Procedural,brewer2017computerized,de2019procedural}: cellular automata, search-based and grammar-based methods, machine learning~\cite{summerville2015learning}, and hybrid methods \cite{liapis2015procedural}. 

%%% I assumed "cooperative" meant "hybrid" here. This extra sentence seemed to unnecesarily focus on one application.

%An example of cooperative methods is \cite{liapis2015procedural}, which used procedural personas for evaluating the playability and quality of generated dungeons levels. 

Mario level generation is a popular test case for PCG methods. Since the 2010 Mario AI Championship~\cite{shaker20112010}, the Mario AI framework has been widely used to validate approaches for generating levels for Super Mario Bros.~\cite{shaker2012evolving,summerville2016super,jain2016autoencoders}. For instance, Summerville and Mateas~\cite{summerville2016super} applied Long Short-Term Memory Recurrent Neural Networks (LSTMs) to generate Mario levels, and then improved the generated levels by incorporating player path information. Jain~\el~\cite{jain2016autoencoders} trained auto-encoders to generate new levels using a binary encoding where empty (accessible) spaces were represented by $0$ and the others by $1$.
More recently, GANs were used to generate Mario levels~\cite{volz:gecco2018,lucas:gecco2019}.

\subsection{Generative Adversarial Networks}

Generative Adversarial Networks (GANs) were first introduced by Goodfellow~\el~\cite{goodfellow2014generative}. Their training process can be seen as a two-player adversarial game in which a generator $G$ (faking samples decoded from a random noise vector) and a discriminator $D$ (distinguishing real/fake samples and outputting 0 or 1) are trained at the same time by playing against each other (Fig.~\ref{fig:architecture}). The discriminator $D$ aims at minimizing the probability of misjudgment, while the generator $G$ aims at maximizing that probability. Thus, the generator is trained to deceive the discriminator by generating samples that are good enough to be classified as genuine. Training ideally reaches a steady state where $G$ reliably generates realistic examples and $D$ is no more accurate than a coin flip. Formally, this training process can be described using the minimax objective:
\begin{equation}
\min_G \max_{\text{D}} \mathop{\mathbb{E}}_{\bm{x}\sim\mathbb{P}_r } [\log ( D(\bm{x}))] + \mathop{\mathbb{E}}_{\bm{\tilde{x}} \sim \mathbb{P}_g} [\log(1-D(\tilde{\bm{x}}))],
\end{equation}
where $\mathbb{P}_r$ is the distribution of the training data and $\mathbb{P}_g$ the distribution generated by  $G(z)$ with $z$ being the input to the generator. While the latent vector $z$ is normally sampled from a noise distribution, in the work presented here it is optimized through interactive evolution.

GANs quickly became popular in some sub-fields of computer vision, such as image generation. However, training GANs is not trivial and often results in unstable models. Many extensions have been proposed, such as Deep Convolutional Generative Adversarial Networks (DCGANs~\cite{radford2015unsupervised}), a class of Convolutional Neural Networks (CNNs); Auto-Encoder Generative Adversarial Networks (AE-GANs~\cite{makhzani2015adversarial}); and Plug and Play Generative Networks (PPGNs~\cite{nguyen2016plug}).
A particularly interesting variation are Wasserstein GANs (WGANs~\cite{arjovsky2017wasserstein,gulrajani2017improved}). WGANs minimize the approximated Earth-Mover (EM) distance (also called Wasserstein metric), which is used to measure how different the trained model distribution and the real distribution are. WGANs have been demonstrated to achieve more stable training than standard GANs.

At the end of training, the discriminator $D$ is discarded, and the generator $G$ is used to produce new, novel outputs that capture the fundamental properties present in the training examples. The input to $G$ is some fixed-length vector from a latent space (usually sampled from a block-uniform or isotropic Gaussian distribution). For a properly trained GAN, randomly sampling vectors from this space should produce outputs that would be misclassified as examples of the target class with equal likelihood to the true examples. However, even if all GAN outputs are perceived as valid members of the target class, there could still be a wide range of meaningful variation within the class that a human designer would want to select between. A means of searching within the real-valued latent vector space of the GAN would allow a human to find members of the target class that satisfy certain requirements.

\subsection{Interactive Evolutionary Computation}

In interactive evolutionary computation (IEC) the traditional objective fitness function employed in evolutionary computation is replaced by a human manually selecting the candidates for the next generation \cite{takagi2001interactive}. IEC has traditionally been used in optimization tasks of subjective criteria or in open-ended domains where it is difficult to define an objective. Another common area of application is where optimization is based on subjective aspects of aesthetics and beauty. All of these characteristics apply to content generation for games. IEC has thus been applied to various areas of content generation for games, as diverse as evolving maps \cite{liapis2013sketchbook,olsted2015interactive}, weapons \cite{hastings2009evolving}, and also flowers \cite{risi2015petalz}. However, none of the current IEC for games employ Latent Variable Evolution, which is described next. 

\subsection{Latent Variable Evolution}
\label{sec:lve}
The first \emph{latent variable evolution} (LVE) approach was proposed by Bontrager~\el~\cite{bontrager2017deepmasterprint}. In their work the authors train a GAN on a set of real fingerprint images and then apply evolutionary search to find latent vectors matching subjects in the dataset. 

In another paper \citeauthor{bontrager2018deep}~\cite{bontrager2018deep} present an interactive evolutionary system, in which users can evolve the latent vectors for a GAN trained on different classes of objects (e.g.\ faces or shoes). Because the GAN is trained on a specific target domain, it becomes a compact and robust genotype-to-phenotype mapping (i.e. most produced phenotypes do resemble valid domain artifacts) and users were able to guide evolution towards images that closely resembled given target images. Such target-based evolution has been shown to be challenging with other indirect encodings \cite{woolley2011deleterious}. In more recent work, \citeauthor{zaltron2019cg}~\cite{zaltron2019cg} extended the approach in \citeauthor{bontrager2018deep} with the ability to freeze and manually edit certain features, allowing users to easily create facial composites.

In video games, LVE has been applied to the first-person shooter Doom using a GAN \cite{giacomello:cog19}. An alternative way to induce a latent space for LVE is via an autoencoder, as done in the game Lode Runner \cite{thakkar:cog2019}. 
This paper builds on the first application of GANs and LVE to video game levels, specifically for Super Mario Bros.\ \cite{volz:gecco2018,lucas:gecco2019,volz:gecco19}. In fact, this paper uses the publicly available GAN models produced from that research. New GAN models for Zelda are also trained using the same code as the Mario models.

%%%%% The PCG section used to come after this one
%%These recent examples are all part of longer history of Procedural Content Generation (PCG) using evolution and/or trained models, which is described in more detail next. 

\section{Video Game Domains}
\label{sec:domains}

The games explored in this paper rely on data from the Video Game Level Corpus (VGLC \cite{summerville:vglc2016}). Specifically, GAN models for Super Mario Bros.\ and The Legend of Zelda are used, though in each case some specialized processing of the data is required.

\subsection{Super Mario Bros.}

Super Mario Bros.~(1985) is a classic platform game that involves moving left to right while running and jumping. In order to visualize and play the levels, the Mario AI framework is used\footnote{http://marioai.org/}. At the start of evaluation, Mario is in his \emph{fire flower} state, which means he can launch fire balls and is the size of two tiles on screen. Contact with an enemy reverts him to \emph{big} state (no more fire balls), and subsequent contact shrinks him to \emph{small} state, which takes up only one tile, meaning that he can access areas that are inaccessible by Mario when large. %Mario has access to more areas on screen \todo{This is not true, Mario can jump higher if bigger, if I am not mistaken}
When Mario is \emph{small}, any further enemy contact causes him to die, ending evaluation.

%% Added discussion about Mario states here because we talk about them later in the study results

The VGLC description of Mario levels is tile-based. 
To use the Mario AI framework,
a mapping from VGLC tiles to Mario AI tiles was required (Table \ref{tab:mariotiles}).
Each Mario level from VGLC uses a particular character symbol to represent each possible tile type. This encoding was extended to represent a wider variety of tile types (e.g.\ different enemies). In order to avoid generating incomplete \emph{pipes}, the VGLC encoding of pipes was modified as well. Instead of using four different tile types for a \emph{pipe}, a single tile is used as an indicator for the presence of a \emph{pipe} and extended automatically as required. A detailed explanation of all modifications made to the encoding can be found in work by Volz \cite[Chap. 4.3.3.2]{Volz19}.

%However, it should be noted that this VGLC representation
%is primarily concerned with functional properties of tiles
%rather than artistic properties, and is thus incapable of
%distinguishing certain visually distinct tile types. The 
%only exception are \emph{pipes}, which are represented by
%four visually distinct tile types, despite all being functionally
%equivalent to an impassable ground block. Interestingly,
%the VGLC encoding ignores functional differences between
%different enemy types by providing only a single character
%symbol to represent enemies, which we choose to map to the
%generic \emph{Goomba} enemy type.
%\todo[inline]{This claim about Goombas is not true in the current encoding, at least not for all models. How did Vanessa change this?}

%\todo[inline]{Vanessa did extra work with Mario models beyond what was in our original GECCO paper. I need her to provide details about how those training sets and models were corrected, including pipe repair.}

\begin{table}[t!]
\centering
\caption{\label{tab:mariotiles}Tile types used in generated Mario levels. \normalfont 
The symbol characters (\textit{sym}) come from the modified VGLC encoding, and the
numeric identity values (\textit{num}) are mapped to the corresponding values
in the Mario AI framework to produce the visualization (\textit{vis}) shown.
Numeric identity values are 
expanded into one-hot vectors when input into the
discriminator network during GAN training.}
\begin{tabular}{lcccl}
\hline
Tile type & sym & num & vis&\\
\hline
Stone & X & 0 & \includegraphics[scale=0.5]{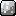}&\\
Breakable & x & 1 & \includegraphics[scale=0.5]{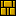}&\\
Empty (passable) & - & 2 & \\
Question Block with coin & q & 3 & \includegraphics[scale=0.5]{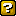}&\\
Question Block with power up & Q & 4 & \includegraphics[scale=0.5]{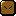}&\\
Coin & o & 5 & \includegraphics[scale=0.5]{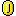}&\\
Pipe & t & 6 & \includegraphics[scale=0.5]{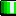}\includegraphics[scale=0.5]{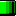}&\\[-6.5px]
& & & \includegraphics[scale=0.5]{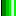}\includegraphics[scale=0.5]{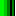}&\\
Piranha Plant Pipe & p & 7 & \includegraphics[scale=0.5]{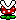}&\\[-6.5px]
 &  &  & \includegraphics[scale=0.5]{encoding_6.png}\includegraphics[scale=0.5]{encoding_7.png}&\\[-6.5px]
& & & \includegraphics[scale=0.5]{encoding_8.png}\includegraphics[scale=0.5]{encoding_9.png}&\\
Bullet Bill & b & 8 & \includegraphics[scale=0.5]{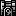}& \hspace{-15px} \includegraphics[scale=0.5]{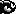}\\[-6.5px]
 &  &  & \includegraphics[scale=0.5]{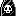}&\\[-6.5px]
 &  &  & \includegraphics[scale=0.5]{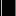}&\\
Goomba & g & 9 & \includegraphics[scale=0.5]{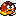} \includegraphics[scale=0.5]{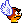}&\\
Green Koopa & k & 10 & \includegraphics[scale=0.5]{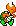} \includegraphics[scale=0.5]{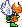} & \\
Red Koopa & r & 11 & \includegraphics[scale=0.5]{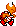} \includegraphics[scale=0.5]{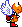} &\\
Spiny & s & 12 & \includegraphics[scale=0.5]{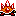} \includegraphics[scale=0.5]{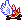} &\\
\hline
\end{tabular}
\end{table}

\subsection{The Legend of Zelda}

The Legend of Zelda (1986) is an action-adventure dungeon crawler. The main character, Link, explores an overworld to find and explore several maze-like dungeons full of enemies, traps, and puzzles. This game also has a tile-based VGLC description of the dungeons (not the overworld). In this paper, the game is visualized and played with a custom-made, ASCII-based Rogue-like game engine\footnote{Based off of \url{http://trystans.blogspot.com/}}, and thus requires a way of mapping the VGLC representation to the Rogue-like representation (Table~\ref{tab:zeldatiles}).

\begin{table}[t!]
\centering
\caption{\label{tab:zeldatiles}Tile types used in generated Zelda rooms. \normalfont
The symbol characters (\emph{sym}) come from the VGLC encoding, and the corresponding
tile images from the original Legend of Zelda are also shown (\emph{game}). %Many tile types (distinct tiles for walls, blocks, and statues) are superfluous in the Rogue-like game engine used to play the evolved levels. Also, doors are not needed because they are placed by the generative graph grammar that makes Zelda dungeons (Section \ref{sec:playLevels}). %%% <-- stated in text
The diversity of VGLC tile types is mapped down to a smaller set of tiles. The numeric values for GAN training are shown (\emph{num}), and the final tiles depict how they appear in the Rogue-like game engine used in this paper (\emph{rogue}).}
\begin{tabular}{ccccc}
\hline
Tile type & sym & game & num & rogue\\
\hline
% Images needed to be vertically centered within row
Floor & F & \includegraphics[scale=0.75]{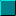} & 0 & \includegraphics[scale=0.75]{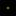} \\
Wall & W & \includegraphics[scale=0.375]{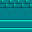} & 1 & \includegraphics[scale=0.75]{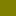} \\
Block & B & \includegraphics[scale=0.75]{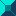} & 1 & 
\includegraphics[scale=0.75]{rouge-wall.png}\\
Door & D & \includegraphics[scale=0.375]{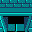} & 1 & \includegraphics[scale=0.75]{rouge-wall.png}\\
Stair & S & \includegraphics[scale=0.75]{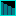} & 1 & \includegraphics[scale=0.75]{rouge-wall.png}\\
Monster statue & M & \includegraphics[scale=0.75]{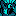} & 1 & \includegraphics[scale=0.75]{rouge-wall.png}\\
Water & P & \includegraphics[scale=0.75]{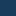} & 2 & 
\includegraphics[scale=0.75]{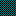}\\
Walk-able Water & O & \includegraphics[scale=0.75]{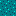} & 2 & \includegraphics[scale=0.75]{rouge-water.png}\\
Water Block & I & \includegraphics[scale=0.75]{block.png} & 2 & \includegraphics[scale=0.75]{rouge-water.png}\\

\hline
\end{tabular}
\end{table}

In this game, the large set of tiles inherent to Zelda was actually reduced to a smaller set based on functional requirements. Some Zelda tiles differ in purely aesthetic ways, and others rely on complicated mechanics not implemented or even necessary in the Rogue-like. 
The Rogue-like accommodates three tile types: a floor tile which directly corresponds to the Zelda floor tile, an impassable tile that corresponds to all impassable tiles in Zelda, and a water tile that corresponds to a semi-passable obstacle. 

In Zelda, water is not passable by Link until he has obtained the raft item, and even then he can only pass over a single water tile. This item is implemented in the Rogue-like, and is in each level. Zelda also has some enemies that can pass over water, and others that cannot. In contrast, all Rogue-like enemies can pass over water.

Enemies are not represented in the VGLC data because the authors neglected to include them\footnote{VGLC erroneously refers to impassable statues that occupy some rooms as enemies, but they are simply impassable objects. Other enemies are absent from VGLC}. In the Rogue-like, they randomly populate certain rooms. VGLC does include information about doors linking rooms, but that information is excluded from the Rogue-like encoding because door placement is handled not by the GAN, but by a generative graph grammar (Section \ref{sec:playLevels}).

\section{Approach}
\label{sec:approach}

Separate GANs are trained on level data from Mario and Zelda. 
Interactive evolution is then used to search the latent spaces
induced by the GAN models. Other interface features are also described. 

\begin{figure}[t]
\centering
\includegraphics[width=\columnwidth]{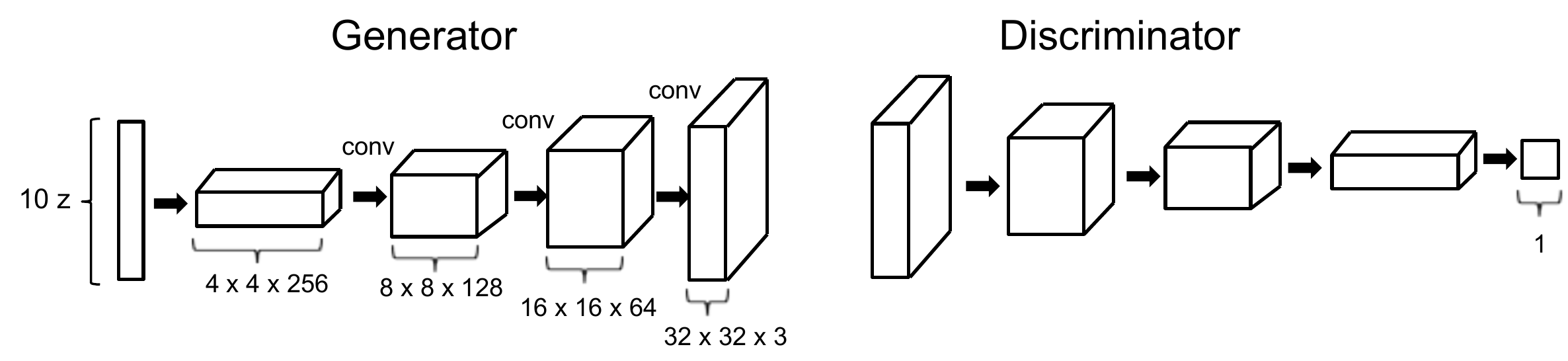}
\caption{GAN Architecture for Zelda. \normalfont Mario uses the same architecture, but with a latent vector size of 5 instead of 10, and an output depth of 13 instead of 3.}
\label{fig:architecture}
\end{figure}

\subsection{GAN Training Details}

The Mario model used in this paper was taken from a publicly
available repository associated with previous research \cite{volz:gecco19}, %\todo{This is not technically true. The GECCO 19 proposal used the old encoding. The new models are only used in my thesis and the games benchmark},
but details of its training are repeated here for clarity.
The Zelda model used in this paper is new, but it is closely
based on another model from a recent paper \cite{gutierrez2020zeldagan}.
Furthermore, the system described in this paper is compatible with all
previously published GAN models associated with these papers.
All models are Wasserstein GANs \cite{arjovsky2017wasserstein} differing
only in the size of their latent vector inputs (10 for Zelda, 5 for Mario), and the depth of the final output layer (3 for Zelda, 13 for Mario). The Zelda
architecture is in Fig.\ \ref{fig:architecture}.
The output depth corresponds to the number of possible tiles for the game. The other output dimensions can match for both games because they are larger than the 2D region needed to render a level segment. During training and generation, the upper left corner of the output region is treated as a generated level, and the rest is ignored.

%\todo{This is unclear. Also: The number of layers is not necessarily the same, correct? Also, my new GAN approach has a different number of layers. Could we not just use variables in the figure to avoid confusion?}

To encode the levels for training,
each tile type is represented by
a distinct integer, which is converted to a one-hot
encoded vector before being input into the
discriminator. 
The generator also outputs
levels represented using 
the one-hot encoded format, which is then converted
back to a collection of integer values.
Mario levels in this integer-based format can be sent to
the Mario AI framework for rendering, and Zelda rooms
in this format can be rendered by the Rogue-like engine.
The mapping from VGLC tile types and symbols, 
to GAN training number codes, and finally to Mario AI/Rogue-like 
tile visualizations
is detailed in Tables~\ref{tab:mariotiles} and~\ref{tab:zeldatiles}. 

The GAN input files for Mario were created by processing all 12 \emph{overworld} level files
from the VGLC for the original Nintendo game \emph{Super Mario Bros}.
Each level file is a plain text file where each line of the file corresponds to a row
of tiles in the Mario level.  Within a level all rows are of the same length,
and each level is 14 tiles high.  The GAN expected to always see a rectangular
image of the same size, hence each input image was generated
by sliding a 28 (wide) $\times$ 14 (high)
window over the raw level from left to right, one tile at a time. The width of 28 tiles is equal to the width of the screen in Mario. 
In the input files each tile type is
represented by a specific character, which was then mapped to a specific integer in the training
images, as listed in Table~\ref{tab:mariotiles}.  

%While we could have used a larger dataset instead of this relatively small one, its use allows us to test the GAN's ability to learn from relatively little data, which could be especially important for games that do not offer such a large training corpus as Mario. Additionally, because of the smaller  training set it is possible to manually inspect if the LVE approach is able to generate levels with properties not directly  found in the training set itself. 

The GAN input for Zelda was created from the 18 dungeon files in VGLC for \emph{The Legend of Zelda}, but the actual training samples are the individual rooms in the dungeons, which are 16 (wide) $\times$ 11 (high) tiles in size. Many rooms are repeated within and across dungeons, so only unique rooms were included in the training set, which consisted of only 38 samples. The training samples are simpler than the raw VGLC rooms because the various tile types are reduced to a set of just three as shown in Table~\ref{tab:zeldatiles}. The tile set was reduced to focus only on the functional aspects of tiles needed for the Rogue-like engine. Additionally, doors were transformed into walls because door placement is not handled by the GAN, but rather by a generative graph grammar which is used to combine separate GAN-generated rooms into playable dungeons (Section~\ref{sec:playLevels}).

%\todo[inline]{Mario details from Vanessa and Zelda details from Jake}
%\todo[inline]{Vanessa: I am not sure what exactly should be going here? The GAN architecture? I kept the same as in the previous paper for Mario}

\begin{figure*}[t!]
\centering
\begin{subfigure}{0.49\textwidth} 
    \includegraphics[width=1.0\textwidth]{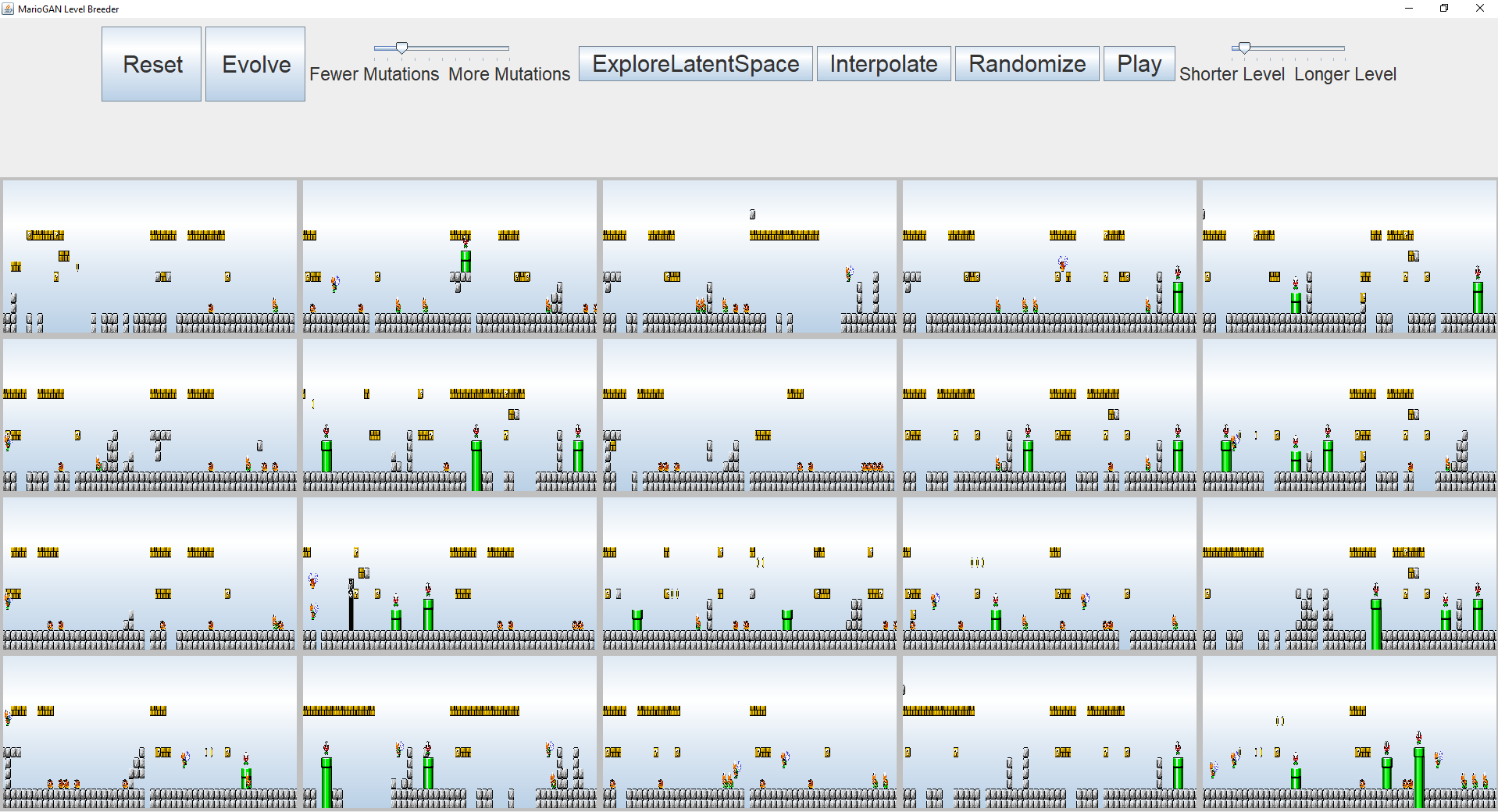}
    \caption{Super Mario Bros.}
    \label{fig:mario}
\end{subfigure}
\begin{subfigure}{0.49\textwidth}
    \includegraphics[width=1.0\textwidth]{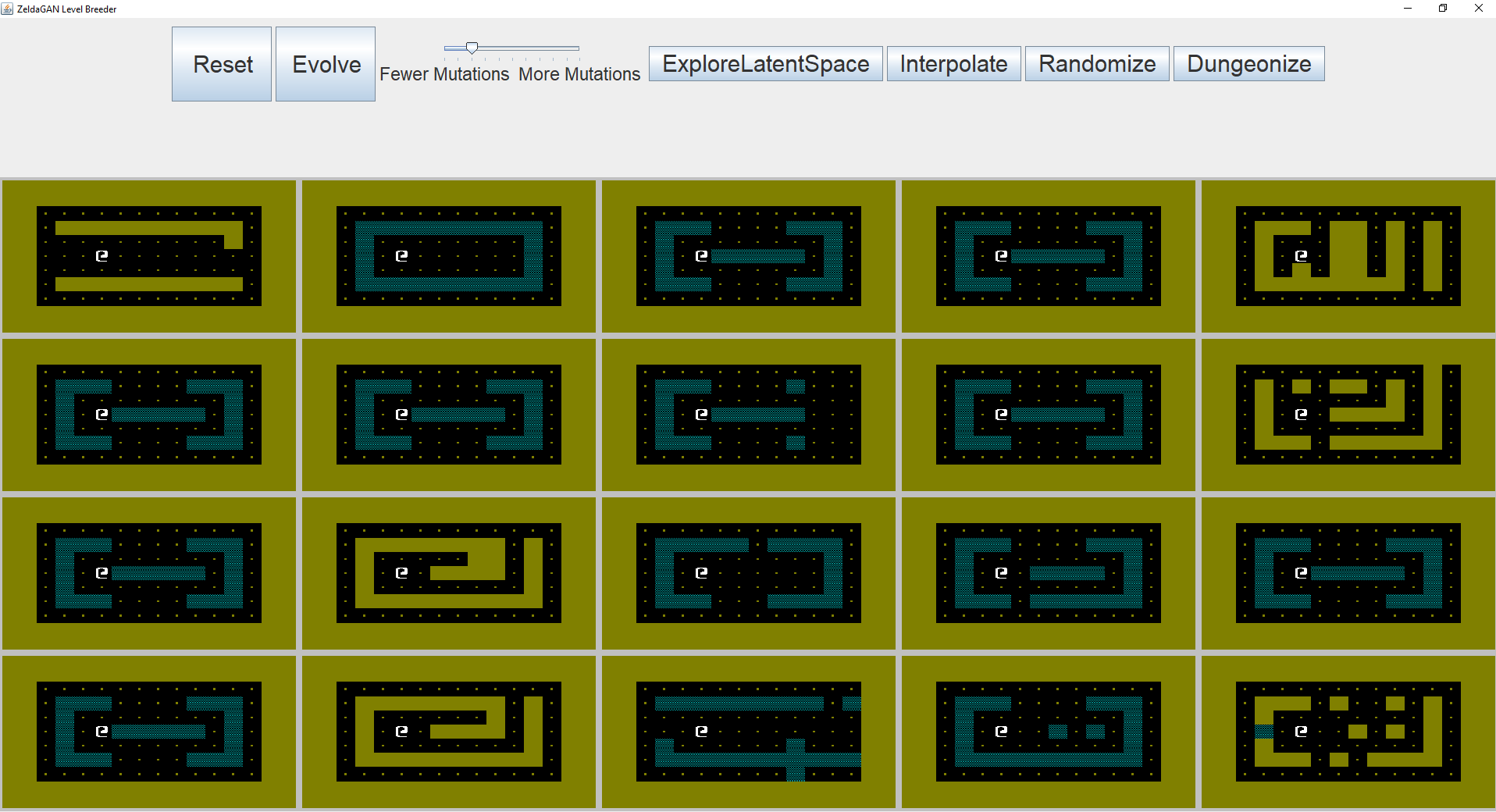}
    \caption{The Legend of Zelda}
    \label{fig:zelda}
\end{subfigure}
\caption{Interactive Evolution Interfaces. \normalfont Levels can be evolved for (\subref{fig:mario}) Mario or (\subref{fig:zelda}) Zelda. Both interfaces display a $5 \times 4$ grid of level sketches. Multiple individuals can be selected before clicking the \emph{Evolve} button to create the next generation. Mario levels can have their length adjusted, and can also be played. Zelda individuals are actually only rooms, but selecting multiple rooms and pressing \emph{Dungeonize} generates a randomized dungeon from the rooms, which can then be played.}
\label{fig:interface}
\end{figure*}

\subsection{Interactive Evolution via Selective Breeding} 

The levels are evolved interactively using an interface similar to Picbreeder \cite{secretan:ecj2011} and other interactive evolution systems (Fig.~\ref{fig:interface}).
A selective breeding algorithm using pure elitist selection is applied. First the user sees $N=20$ images representing level sketches. Each level sketch is generated by a latent vector with a length dependent on the particular GAN model used to generate it. Complete Mario levels are depicted with the tile graphics from Table \ref{tab:mariotiles}, and individual Zelda dungeon rooms are depicted with the tiles from Table \ref{tab:zeldatiles}.

The user selects $M<N$ individuals as parents for the next generation. These parents are directly copied to the next generation, and remaining slots are filled by their offspring. 
Selection continues in this fashion for as long as the user likes. 
There is a 50\% chance of crossover for each offspring (single-point crossover on the vectors of real numbers). 
Independent of whether an offspring has two parents or is a clone, it then has a certain number of chances to mutate defined by a user-controlled slider ranging from 1 to 10. For each mutation chance, each real-valued number in the vector has an independent 30\% chance of polynomial mutation \cite{deb1:cs95:polynomial}. %\todo{I don't really get it? So you apply polynomial mutation $m$ times, where $m$ is set by the slider?} 
The slider for the number of mutation chances allows users to control the amount of exploration done, by expressing a preference for few or many changes to the genome.

Sometimes, evolution can converge and get stuck in an area of the search space that is hard to escape \cite{goldberg1987simple}. This problem is especially common with small populations. One way around premature convergence is to simply restart the evolution process and hope for better results. Therefore, this system provides users with the option to replace selected genomes with new random ones via the \emph{Randomize} button. The fact that only selected individuals are replaced allows users to maintain the best levels they've discovered so far while also looking for potential new starting points for a search.
However, users can also explore the search space by changing vectors in other ways.

\subsection{Manual Exploration of Latent Space} 

\begin{figure}
    \centering
    \includegraphics[width=\columnwidth]{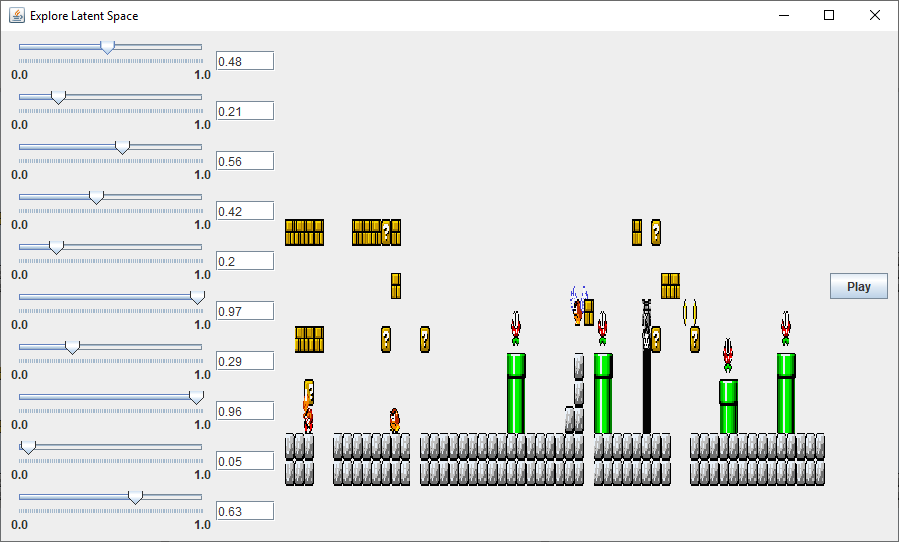}
    \caption{Interface for Manual Exploration of Latent Space.  \normalfont     Direct modification of the 10 latent vector values on the left immediately updates the visualization of the Mario level, which can be played by the provided button. The Mario level has 10 latent variables because it consists of two segments, where each is generated by a latent vector of length 5.}
    \label{fig:explorelatent}
\end{figure}
Users can explore the induced latent vector space more directly with the \emph{Explore Latent Space} option (Fig.~\ref{fig:explorelatent}). This button can be clicked after selecting an individual from the population. A new window pops up with a visualization of the selected level and a number slider for each real number in the genotype. Next to each slider is an input box showing the currently selected value.

Users can change any individual real-valued component of the genotype by either manipulating the slider or entering a new number in the adjacent input box. The visualization updates accordingly. Users can even select a slider and use the left and right arrow keys to watch how the visualization of the level morphs with the changing inputs. Individual tiles gradually swap with others in a relatively smooth fashion, given that every location can only contain one tile. These sliders allow a user to explore the area of latent space surrounding a particular individual. Any modifications during exploration affect the genotype of the individual, thus allowing the modified vector to be selected for further evolution.

\begin{figure}
    \centering
    \includegraphics[width=\columnwidth]{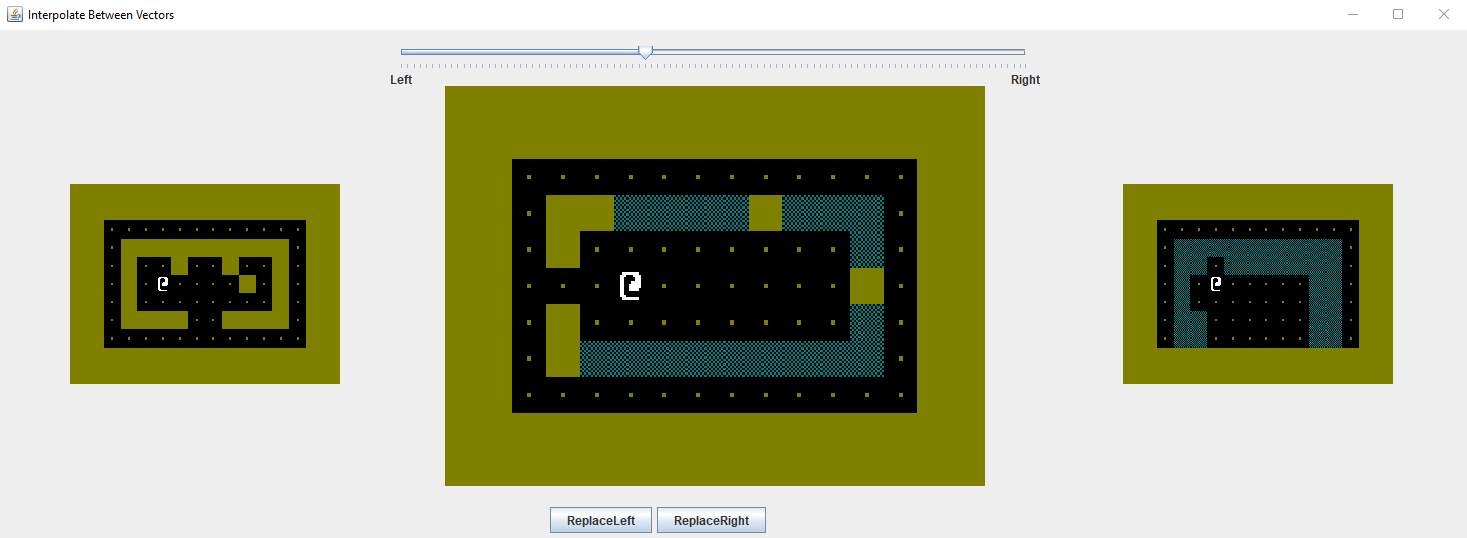}
    \caption{Interface for Interpolation Between Latent Vectors.  \normalfont   The slider at the top adjusts the large Zelda room
    in the middle to be at different positions along the line in latent space connecting the latent vectors for the rooms
    on the left and right.}
    \label{fig:interpolate}
\end{figure}

Another way to explore the latent space is via the \emph{Interpolate} button. This option requires two individuals to be selected, and opens a window containing one level on the left, and the other on the right. Between the levels is another visualization with a slider above it. The slider determines to what extent the center level is closer in latent space to either the left or right level (Fig.~\ref{fig:interpolate}).
Specifically, the slider traverses a line in high-dimensional space connecting the left level's vector to the right level's vector. Each increment along the slider moves each individual numerical component of the center vector closer to the corresponding component in either the left or right vector. The center level updates as the slider moves, and allows users to find a level that blends aspects of two other levels. In order to bring the newly created level into the population, it must replace either the left or right extreme, and the interface has buttons allowing either of the two levels to be replaced.

Exploring the search space with interpolation and direct modification of individual values allows for finer control of the generated levels, but the various latent dimensions seldom have a direct relation to particular level components. Simply picking what looks appealing and evolving it can sometimes be more straightforward.
However the space is searched, a way to play the levels is needed to properly evaluate them.

\subsection{Interaction With Generated Content}

\label{sec:playLevels}

Users can interact with content generated for each game.

Because a single latent vector represents a whole Mario level, the user can select a level and then click \emph{Play} to launch the level in the Mario AI framework. Users can also modify the number of segments in the level. If the number is increased, then the underlying latent vector genome is extended by repeatedly copying the original latent vector values. Further evolution or direct manipulation can differentiate these sub-vectors from each other. To convert a lengthened vector into a level, it is split into sub-vectors of the appropriate length, which are sent to the GAN individually to create segments that are concatenated into one big level.

\begin{figure}
    \centering
    \includegraphics[width=\columnwidth]{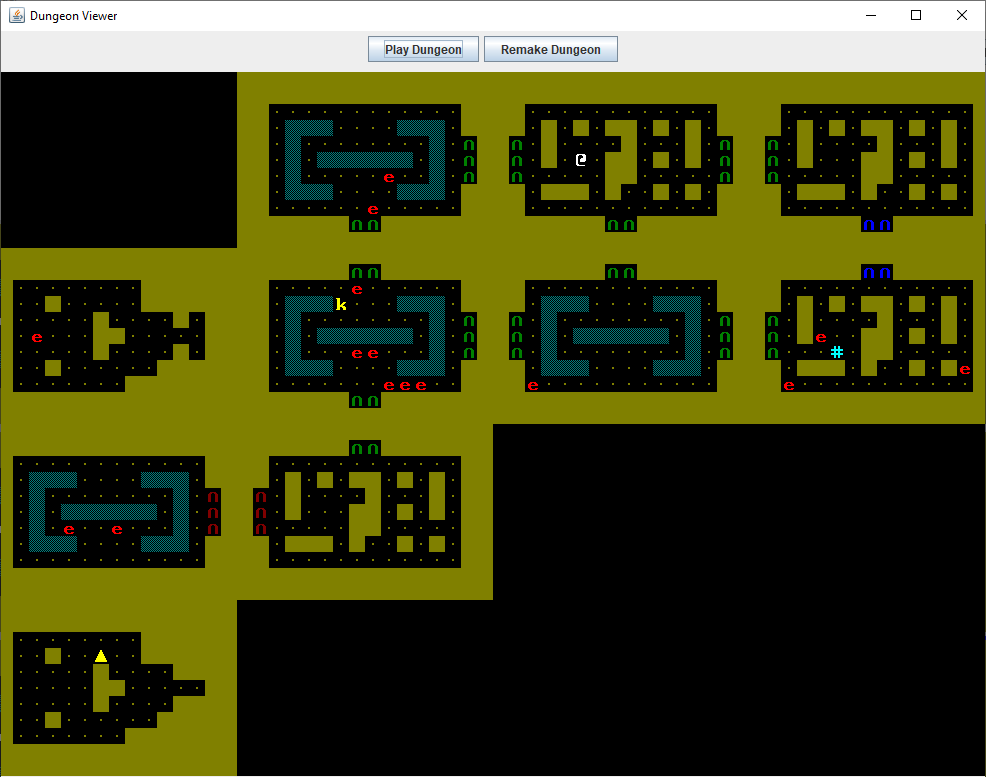}
    \caption{Dungeon Generated By Graph Grammar From GAN Rooms.  \normalfont   This dungeon was generated by selecting three rooms and clicking \emph{Dungeonize}. Its layout is random, but based on a graph grammar. The user-selected rooms are randomly placed. There are several symbols: @ = player avatar, e = enemy, k = key, \#~=~raft, and $\triangle$ = Triforce/goal. Doors can be green (normal), red (locked), or blue (opens after enemies are defeated). Some of the walls are bombable, but these are not marked. The user can either play this dungeon, or choose to generate a new one using the same rooms.}
    \label{fig:dungeonview}
\end{figure}

Latent vectors for Zelda models represent individual rooms, but Zelda dungeons consist of multiple rooms. Instead of a \emph{Play} button, there is a \emph{Dungeonize} button. The user first selects several rooms, and then clicks \emph{Dungeonize} to see a preview of a dungeon generated from the selected rooms (Fig.~\ref{fig:dungeonview}). These dungeons are generated by a graph grammar \cite{dormans:pcg10}. Full details of how to combine a GAN with a graph grammar to generate dungeon levels for Zelda are discussed elsewhere \cite{gutierrez2020zeldagan},
but some key points are mentioned here.

A graph grammar takes an initial \emph{backbone} graph containing non-terminal symbols. An iterative process gradually replaces each symbol with sub-graphs until a graph with only terminal symbols remains. Each terminal symbol represents an individual room, and the rules of the grammar place various types of interesting challenges in the rooms: enemies, locked doors, keys for the locks, puzzle rooms, etc. This process also determines whether rooms are connected by a door, and the type of the door (normal, locked, puzzle-locked). However, the grammar does not define the layout of the individual rooms.

When mapping the final graph of terminal symbols to a 2D layout of rooms, specific rooms are randomly chosen from the group selected by the user before clicking \emph{Dungeonize}. The number of rooms in the dungeon is determined by the graph grammar and random chance, so user-selected rooms may appear multiple times, or not at all (unlikely unless a large number of rooms is selected). If the user does not like the resulting dungeon, then clicking the \emph{Remake Dungeon} button will generate a new dungeon with the same user-selected rooms.
Generated dungeons can be played using a game engine for Rogue-like style games that uses simple ASCII graphics. This Rogue-like engine is also described elsewhere \cite{gutierrez2020zeldagan}.

\section{Human Subject Study}
\label{sec:humanStudy}

A small user study was conducted to gauge user preference for interactive evolution vs.\ direct latent space exploration. User quotes also give insight into the strengths and weaknesses of the system.

\subsection{Procedure}

The software was tested by convenience sampling 22 individuals 
%undergraduate and graduate students \todo{Is this accurate of modl AI people? Should we just say "22 individuals" or something in between? Seb: Some were native english speakers, so maybe better to remove that part.} 
in China and Denmark
with varying game-playing experience. 
Participants were instructed to design the ``coolest''  Mario/Zelda level possible using the available options. Each user had their own subjective definition of what ``cool'' means.
Participants were invited to evaluate three variations of the tool using the procedure below:
\begin{enumerate}
    \item (design and play) Design levels using only evolution and randomize options. (5 minutes)
    \item (design and play) Design levels using only exploration of latent space, interpolate, and randomize options. (5 minutes)
    \item (survey) Indicate which method led to cooler levels, and explain why.
    \item (design and play) Design levels with all options: evolution, exploration, interpolate, and randomize. (5 minutes)
    \item (survey) Indicate if combining evolution, explore, interpolate, and randomize leads to even better levels, and explain why.
\end{enumerate}

To avoid ordering effects, the order of the first two sessions alternated for different subjects. This study gives insight into what the system is capable of, and which aspects of it should be further developed in the future.

\subsection{User Responses}

\setlength\tabcolsep{2pt}
\begin{table}[t]

\begin{center}
\centering
\caption{\label{tab:userStudy}Human Subject Study Results. \normalfont 
The top two rows show preferences in the first survey, after using the evolution and exploration systems. Note that one Mario user neglected to answer this question. The lower three rows show user perception of the combined system. The main result is that the study participants sligthly preferred the exploration interface over the pure evolution interface and preferred the combination of both the most.}
\begin{tabular}{|r|c|c|c|}
\hline
                    & Mario       & Zelda       & Total       \\ \hline 
\hline 
Prefers Evolution   & 3 (33.33\%) & 5 (41.67\%) & 8 (38.10\%) \\ \hline
Prefers Exploration & \textbf{6} (66.67\%) & \textbf{7} (58.33\%) & \textbf{13} (61.90\%)\\ \hline
%Total               & 9           & 12          & 21          \\ \hline
\hline
Combination Worse   & 1 (10.00\%) & 0 (0.00\%)  & 1 (4.55\%)  \\ \hline
Combination Equal   & 3 (30.00\%) & 4 (33.33\%) & 7 (31.82\%) \\ \hline
Combination Better  & \textbf{6} (60.00\%) & \textbf{8} (66.67\%) & \textbf{14} (63.64\%)\\ \hline
%Total               & 10          & 12          & 22    \\ \hline

  \end{tabular}
\end{center}

\end{table}

Participant responses are summarized in Table \ref{tab:userStudy}. Sample sizes are too small to draw statistically significant conclusions, but results are consistent across the two games. There is a small preference for direct exploration of latent vectors over evolution, but a majority also believes that the combination of approaches is better than either one in isolation. This preference is not surprising, given that extra features generally should not detract from the usefulness of the system.

Only one individual produced worse levels with the combined system than with their individual preference, which was direct exploration. Those that thought the combined system was only as good as their individual favorite seemed to be primarily focusing on the approach they liked more. One user that preferred evolution said, ``I ended up focusing on evolution and randomization only [because] I was not aware how changing the numbers [could] improve my level and clicking two buttons (evolve, random) somehow [gave] me interesting levels.'' Recall that the \emph{Randomize} button was available in all interfaces. A user that preferred exploration stated, ``I didn't feel [that] evolution added much, and the explore latent space [option] was the more interesting method. [Combining methods] didn't negatively affect the process, but it wasn't better than the previous version.''

However, some individuals found that combining evolution and exploration opened up new possibilities, and gave quotes like, ``I was able to bootstrap the characteristics I wanted the level to have by fine-tuning two [levels'] latent vectors. After that, evolution `was aware' that I wanted e.g.\ more green pipes,'' and ``It takes the advantages of the previous two [systems] and it added the [functionality] of generating better levels. The `evolve' and `interpolate' [buttons] help to discover better levels.'' These quotes indicate that a sophisticated user can take full advantage of the system to produce better results, but other quotes indicate that combining these features can be challenging: ``It [requires] skill to combine [these] two [methods]. [It] is difficult.''

%\todo[inline]{Should we correct grammar errors in quotes using brackets, leave the errors in and add (sp) to indicate the grammar errors, or maybe just make a note somewhere that many participants are non-native speakers (already hinted at by the different cultures comment above)? Seb: I don't think it's necessary to correct them and also removed the non-native speaker part since that seemed not relevant for the study.}

\begin{figure}
    \centering
    \includegraphics[width=.8\columnwidth]{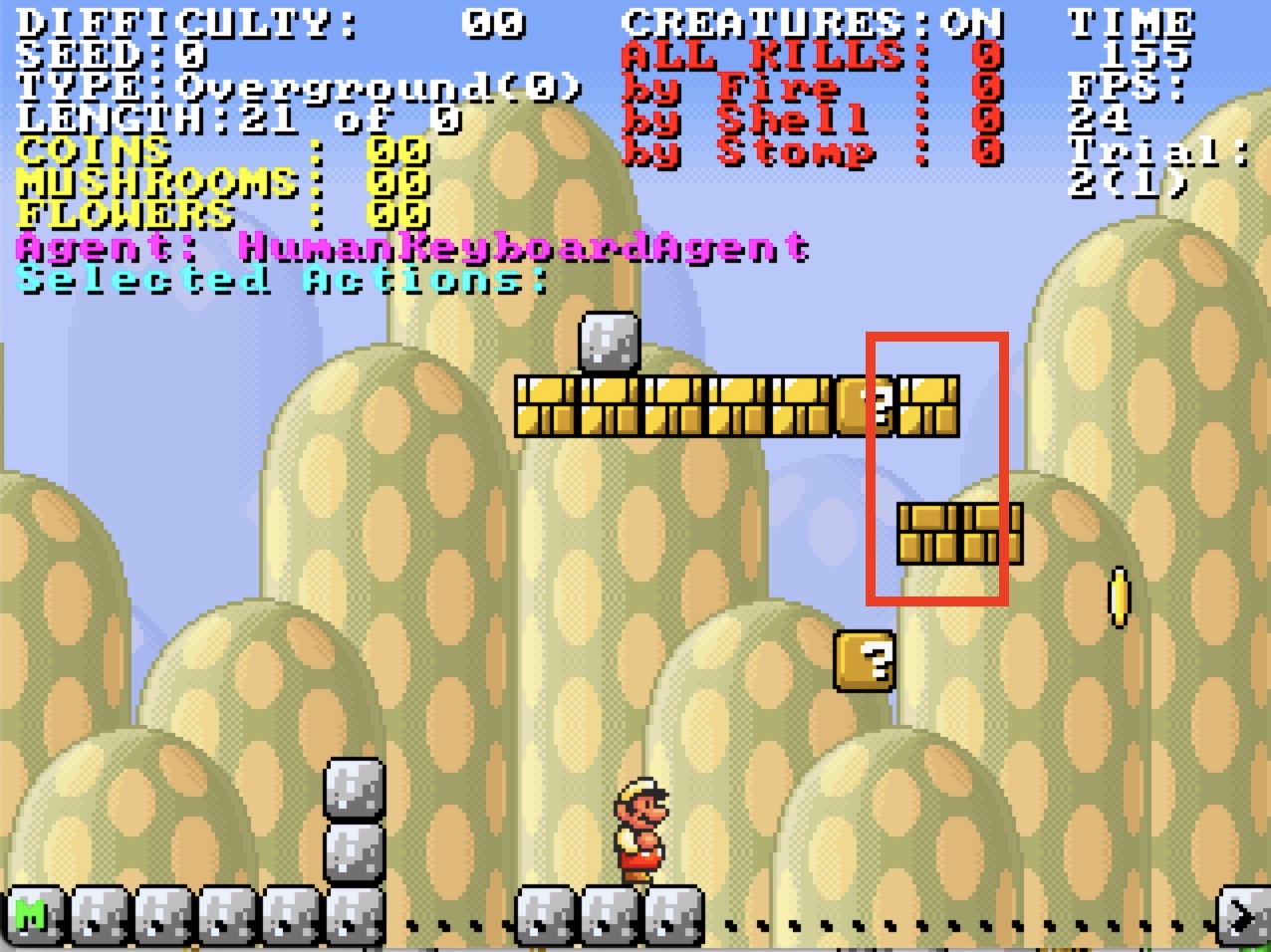}
    \caption{Mario Level Generated By Study Participant. \normalfont To pass the area marked by the red rectangle, Mario needs to be in his \emph{small} state, which the user found to be an unusual limitation.}
    \label{fig:hardmario}
\end{figure}

Some user complaints had more to do with the GAN or other aspects of level generation than with the particular exploration method. Users were surprised that some levels were simply not beatable, or perhaps not beatable without unusual behavior.
For example, in some of the generated Super Mario levels, the only way to pass is to injure Mario so that he shrinks, allowing Mario to pass through tight spaces (Fig.~\ref{fig:hardmario}). A similar issue that affected some Zelda levels is that enemies and items will occasionally appear in unreachable areas of rooms. Even when such placement does not affect whether the level is beatable, users typically find it unusual.
Users also noted that isolated water or wall tiles could be generated, which did not seem very useful. However, some rooms in the original game are empty except for two standalone blocks, so such generated rooms do not differ greatly from levels in the original game.

%\todo[inline]{Jialin to add more about the effect of agent testing here, which may lead to the hard levels or be able to pass such levels}

%In the Zelda dungeons, the selected rooms can be re-used multiple times with different numbers of enemies which lead to different difficulty levels of the game and an increase of playability reported by the human testers. 

Many participants commented that it would be more explainable and easier to design levels if the effects of changes to the latent vector were correlated with particular level features. One said, ``The [system] using explore latent space is not explainable and quite random when changing the values of [a] latent vector.'' Another said, ``It was, however, quite hard to go in a direction, like `I want more enemies'.'' The tile-based nature of the GAN output probably makes working toward a specific goal difficult. Enemies and other tiles suddenly appear from nothing as latent value sliders are manipulated.

%\todo{We could add something about the fact that this type of exploration is made more difficult by the fact we have discrete phenotypes}

%\todo{I can't find quotes for this ... I assume this was through verbal communication}
Some of the participants expected the functionality of direct editing of blocks, via selecting a tile and then changing it to an object (e.g.\ brick or empty) to fix or improve a generated level. For instance, this functionality would be useful to fix the isolated wall issue mentioned previously.

Despite these minor complaints, the users seemed to enjoy the experience. Participants described their levels with words such as ``interesting,'' ``exciting,'' ``diversity,'' and ``innovation.'' The fact that they were able to discover such levels within short five-minute sessions indicates that the GAN helped them focus on levels within a high-quality region of design space.

%Because users found some of the generated content odd, it is useful to evaluate how novel the user-evolved levels are in comparison with the original levels.

%\subsection{Novelty of Evolved Content}

%\todo[inline]{Simon's analysis using JSD can go here.}

\section{Discussion and Future Work}
\label{sec:discussion}

The combination of interactive evolution with latent variable evolution makes it easy to explore a space of high-quality solutions. Furthermore, the discrete space and small size of the tile-based levels makes use of small latent vectors reasonable. In turn, this makes exploration via direct manipulation of latent vector values manageable. Exploring latent space in this way is particularly useful for video game levels, because a user's personal preferences may not be easily quantifiable as an objective function. In fact, in a single session a user's goals may shift as different parts of the latent space are explored. This latent exploration combined with the ability to play the levels to test them out makes the search process itself an entertaining game experience.

Although this interactive search tool allows users to gain a better understanding of latent level design space, it lacks some practical features that would make it more useful to actual level designers. The most obvious feature to add is the ability to directly edit levels by assigning particular tiles to particular locations in a level. If this feature were added, then designers could explore latent space until they stumbled across a level that is close to what they want, and then make small tweaks to perfect it.

Such direct edits to the level could even be incorporated back into the genome. One option to do this is by using a hybrid encoding \cite{helms:pone2017}: genomes would consist of both a latent vector (indirect encoding) and a vector with one slot for every tile in the level (direct encoding). The direct part could act as a filter, overriding the GAN output where specified. These edits could remain fixed, or also be subject to further evolution. Instead of a hybrid encoding, explicit tile edits could also be recorded by searching for a latent vector that matches the edited level (as closely as possible). 
%By default, the direct portion would be initialized to not edit the GAN output. Once the user has edited the directly encoded portion, it would override the GAN output. These edits could remain fixed, or also be subject to further evolution.

Another adjustment that would make the tool more usable is if the latent dimensions actually corresponded to level features meaningful to a human user. Although adjusting the latent dimension sliders results in small local changes, those changes can introduce totally unexpected features, such as changing a staircase of blocks into pipes in Mario, or a maze of water into a thick wall in Zelda. If specific sliders affected specific portions of the level, or directly corresponded to the presence of certain features (e.g.\ number of pipes, denseness of water tiles), then humans could make easy use of them. Such control over latent space could be possible if the GAN had either a special architecture or training regime that made certain latent dimensions more responsible for training loss associated with certain output regions or tile types.
In addition, it could be useful to integrate related work on scaling the response of control parameters for content generation in order to relate better to user expectations \cite{Cook2019}.

Another adjustment to the tool that could improve usefulness would be to extend it towards a mixed-initiative system \cite{liapis2013sketchbook} or a system in which  IEC is combined with novelty search \cite{lehman:ecj2011} and  objective-based search \cite{woolley2014novel,lowe2016accelerating}.
%\todo{Cite something here} by allowing users to occasionally evolve toward objectives, and then interactively select from the results of objective evolution. 
This approach could allow users to quickly move toward different regions of the search space without having to manually lead the system there.

%\todo[inline]{Vanessa: talk about exploring the latent space, and discrete level structure}

%\todo[inline]{Latent vector can be short, but still effective}

%\todo[inline]{Point: hard to think of an objective. Most latent results are decent, and then human finds what they want the most}

%\todo[inline]{Point: Interactive evolution very expensive, but GAN starts at a place where most levels are viable ... saves time.}

Adding most of these practical features to this particular system should be straightforward. Once the system is enhanced, a larger human subject study with a broader sample of participants can be conducted to see if human evaluation reveals statistically significant impacts from particular interface components.

Further development of this system also makes sense because of how general it is. The exact same GAN code was used to train models for both Mario and Zelda. The only game-specific knowledge that the system's code needs is awareness of the output dimension for individual training samples, and a way to display and play the generated levels. Therefore, the authors hope that data for other games will be used to train GAN models and evolve new levels using this system.

%%% Partially addressed at end of previous section
%\todo[inline]{Do we have any way of figuring out whether our initial claim that we have user fatigue when we start from GANs is correct? Do we at least have no one complaining of being bored? or can we visually confirm that the levels look like levels? I feel at least something should be discussed here, as this is the main added value of the paper.}

An additional benefit of providing a human-in-the-loop method of exploring the latent space of GAN-generated content is the great potential of additional insights into the genotype-to-phenotype mapping encoded in GANs. This is an under-explored area, in particular for phenotypes with direct encodings. Data from user-interaction could thus help develop new, better suitable optimization algorithms for these types of problems \cite{volz:gecco19, Volz19}.

\section{Conclusions}
\label{sec:conclusion}

As Generative Adversarial Networks (GANs) continue to be applied as an indirect genotype-to-phenotype mapping
for evolution, practitioners will need tools for effectively exploring the inscrutable latent spaces they produce.
This paper presents a tool for applying interactive latent variable evolution to GAN models that produce video game levels. This allows users to quickly explore level design space  without any a priori objectives in mind.
%which allows users to quickly explore level design space space without any a priori objectives in mind. 
The tool also provides users with more direct ways of controlling the output, namely the features to directly edit latent variables, and to interpolate between points in latent space.
The tool is evaluated based on its application to level generation 
for Super Mario Bros.\ and The Legend of Zelda.
A majority of participants in a human subject study found that the combination of these features was more powerful than
either one in isolation, and gave feedback that will allow this system to be developed further into a general purpose tool
for game level design for other video games.

%that have a tile-based level representation.\todo{what makes this restrict to tile-based levels? I don't see any reason this could not be applied to anything generated with GANs?}

%%% Put back for camera ready
%\begin{acks}

%The authors would like to thank the Schloss Dagstuhl team and
%the organisers of Dagstuhl Seminars 17471 and 19511 
%for hosting productive seminars.

%\end{acks}

\begin{acks}
%%% Will comment from submission, but wanted to have an idea of space

The authors would like to thank 
the Schloss Dagstuhl team and
the organisers of Dagstuhl Seminars 17471 and 19511 
for hosting productive seminars.

\end{acks}

\bibliographystyle{ACM-Reference-Format}
\bibliography{GANExplorer} 

\end{document}